\pdfoutput=1

\documentclass[11pt]{article}

\usepackage[]{acl}

\usepackage{times}
\usepackage{latexsym}

\usepackage[T1]{fontenc}

\usepackage[utf8]{inputenc}

\usepackage{microtype}

\usepackage{inconsolata}
\usepackage{booktabs}
\usepackage{amssymb}
\usepackage{pifont}
\newcommand{\cmark}{\ding{51}}
\newcommand{\xmark}{\ding{55}}
\usepackage{setspace}
\usepackage{enumitem}
\usepackage{xcolor, soul}
\sethlcolor{yellow}
\usepackage{graphicx}
\usepackage{multirow}
\usepackage{csquotes}

\newcommand{\dataset}{\textsc{CARE}}

\title{\dataset{}: Extracting Experimental Findings From Clinical Literature}

\author{Aakanksha Naik$^1$ \quad \quad Bailey Kuehl$^1$ \quad \quad Erin Bransom$^1$\\\textbf{Doug Downey}$^1$ \quad \quad \textbf{Tom Hope}$^{1,2}$ \\ \\
$^1$Allen Institute of AI, USA \\
$^2$The Hebrew University of Jerusalem
}

\begin{document}
\maketitle
\begin{abstract}
Extracting fine-grained experimental findings from literature can provide dramatic utility for scientific applications. Prior work has developed annotation schemas and datasets for limited aspects of this problem, failing to capture the real-world complexity and nuance required. Focusing on biomedicine, this work presents \dataset{}---a new IE dataset for the task of extracting clinical findings. We develop a new annotation schema capturing fine-grained findings as n-ary relations between entities and attributes, which unifies phenomena challenging for current IE systems such as discontinuous entity spans, nested relations, variable arity n-ary relations and numeric results in a single schema. We collect extensive annotations for 700 abstracts from two sources: clinical trials and case reports. We also demonstrate the generalizability of our schema to the computer science and materials science domains. We benchmark state-of-the-art IE systems on \dataset{}, 
showing that even models such as GPT4 struggle. We release our resources\footnote{CARE is available at \url{https://github.com/allenai/CARE}}
to advance research on extracting and aggregating literature findings.   



\end{abstract}

\section{Introduction}

\blockquote[Archie Cochrane, 1979][]{It is surely a great criticism of our profession that we have not organised a critical summary, by specialty or subspecialty, adapted periodically, of all relevant randomised controlled trials.}

Though this critique focused on clinical trials, the statement arguably applies to much of science today. There is tremendous potential utility in extracting, structuring and aggregating fine-grained information about experimental findings and the conditions under which they were achieved, across scientific studies. Once extracted and aggregated, scientific findings can power many critical applications such as producing literature reviews \cite{deyoung-etal-2021-ms}, supporting evidence-based decision-making \cite{naik-etal-2022-literature}, and generating new hypotheses \cite{wang2023learning}. 

\begin{figure}
\centering
\includegraphics[scale=0.53]{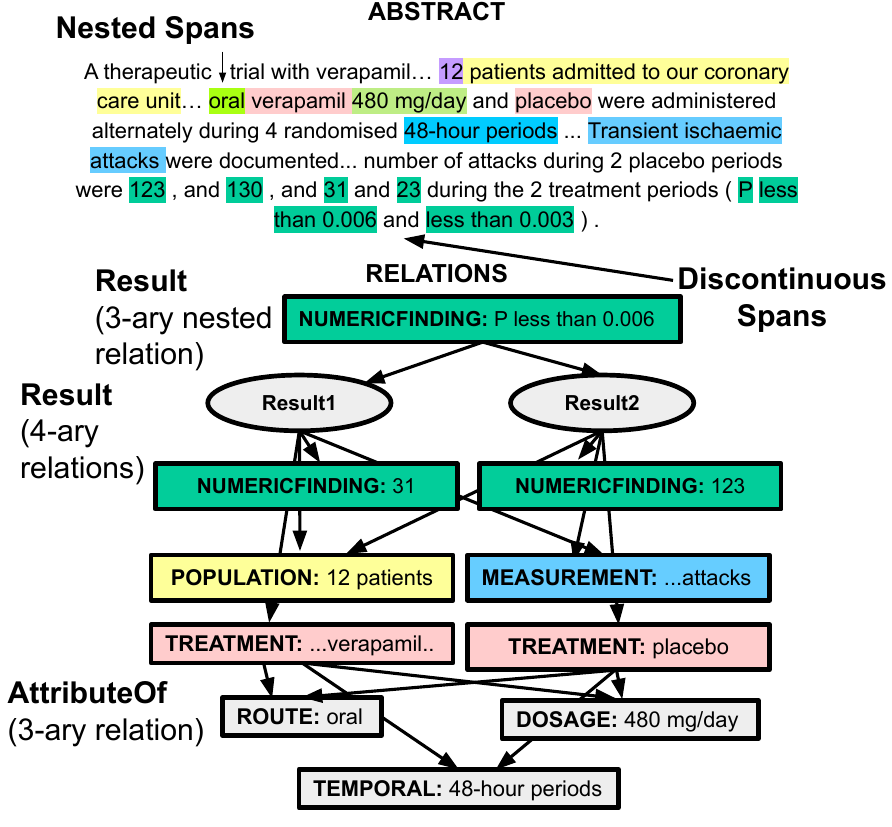}
\caption{A partial example of entity, attribute and relation annotation using our schema for a clinical trial.} 
\label{fig:trialanno}
\end{figure}

While there have been 
efforts on building resources and tools to capture findings in various domains such as clinical trials \cite{lehman2019inferring}, computer science \cite{jain-etal-2020-scirex} and social and behavioral sciences \cite{magnusson2021extracting}---a major obstacle has been 
creating a representation that is \emph{expressive} enough to capture complex and nuanced information about findings. We propose a new representation schema that makes important progress in capturing the real-world complexity of scientific findings in papers, and use it to build a high-quality annotated dataset focusing on biomedical (clinical) findings. Our schema represents fine-grained information about experimental findings and conditions as n-ary relations between entities and attributes, and includes several structural complexities such as discontinuous span annotation, variable arity in relations and nestedness in relations. These aspects have been studied individually in previous datasets \cite{karimi2015cadec,tiktinsky2022dataset}, but our schema is the first to unify them. Our dataset also captures \emph{numeric} findings in addition to their interpretation (e.g., significance, utility, etc.); prior datasets  typically focus solely on the latter (e.g., \citet{lehman2019inferring} captures \emph{increases/decreases} in outcomes but not their magnitudes). 

Though these factors make our annotation schema more complex than prior work, the additional nuance it affords can power several high-impact downstream use cases. For instance, the process of conducting a systematic review\footnote{\url{https://training.cochrane.org/handbook/current}} includes extracting data about specific outcomes from a large number of relevant studies. Unlike prior schemas which only capture interpretation and not precise numeric outcomes, extractors trained on our schema can extract relevant data to assist the review process. Moreover, fine-grained population and treatment details captured by our schema makes it capable of answering highly complex clinical questions (e.g., ``Does vaccination significantly improve mortality outcomes for female patients over 65 who required ventilation?''). Therefore, models trained on our schema can assist physicians in developing more personalized treatment plans, especially for patients with multiple comorbidities or health conditions. Our schema can also help with the development and exploration of richer clinical hypotheses due to its additional granularity.

To build our dataset, named \dataset{} (Clinical Aggregation-oriented Result Extraction), we collect extensive annotations for 700 abstracts (clinical trials and case reports). We also conduct annotation studies demonstrating that our schema generalizes to computer science and materials science, using minor updates based on analogies between aspects across experimental domains (e.g., \emph{populations/interventions} $\rightarrow$ \emph{tasks/methods} in CS). This reflects the expressive power of our schema to generalize across domains while capturing granular and useful information, making it a strong "backbone schema" for research efforts on result-oriented scientific IE. 

We achieve good agreement scores (0.74-0.78 partial F1) comparable to prior work that used simpler schemas that are easier to annotate \cite{luan-etal-2018-multi,nye2018corpus}, and at the same time our resulting dataset is 
larger in size than previous corpora. Our final dataset annotation is extremely rich; at 16.23 relations per abstract, our relation density is nearly 4x that of prior work on annotating findings from clinical trials \cite{lehman2019inferring}. 



We evaluate a wide range of IE models on our dataset, including both extractive systems and generative LLMs. Given the high annotation burden, we test generative LLMs in both fully supervised as well as zero-shot and few-shot settings. Our results demonstrate the difficulty of our dataset, with even SOTA models such as GPT4 struggling to accurately extract clinical findings. As a highly challenging new dataset designed to be reflective of real-world nuance and informational needs, we hope CARE\footnote{\url{https://github.com/allenai/CARE}} is an important resource for the scientific NLP and IE research community to pursue.

\section{Related Work}

\subsection{Information Extraction from Scientific Literature}
Much prior work has focused on information extraction from scientific papers \cite{luan-etal-2018-multi,jain-etal-2020-scirex}, including biomedical literature (see \cite{luo2022biored} for a detailed summary). Most relevant to our goal in this work is prior research on extracting findings or results from scientific literature, which has only explored limited aspects of this problem. 

\citet{gabor-etal-2018-semeval} and \citet{luan-etal-2018-multi} annotate \emph{associative} relations between entities being compared or producing a result, as part of their broader goal of developing IE resources for computer science, but do not capture any nuance (e.g., directionality, causality, etc. of results). Conversely, \citet{magnusson2021extracting} develop a schema focused solely on capturing associations between experimental variables and evidence. However, their focus on sentence-level annotation from scientific claims limits how much additional nuance about experimental setting can be captured.

\begin{table*}[]
    \centering
    \small
    \begin{tabular}{lccp{10cm}}
   \toprule
    \textbf{Type} & \textbf{EBM} & \textbf{CTKG} & \textbf{Example} \\ \midrule
    \textbf{Population} & \cmark & \cmark & This study compared rizatriptan 5 mg and placebo in \hl{1268 outpatients treating a single migraine attack} \\ 
    \textbf{Subpopulation} & \cmark & \cmark & We found low-certainty evidence of little or no difference in delirium (RR 1.06, 95\% CI 0.55 to 2.06; 2 studies, \hl{800 participants})\\ 
    \textbf{Treatment} & \cmark & \cmark & \hl{Dialysate magnesium} was 0.375 mM/L for the hemodialysis\\ 
    \textbf{Measurement} & \cmark & \cmark & \hl{Headache relief rates} after rizatriptan 10 mg were higher\\ 
    \textbf{Temporal} & \xmark & \cmark & After a \hl{48-hour run-in period} , oral verapamil 480 mg/day and placebo were administered\\ 
    \textbf{NumericFinding} & \xmark & \cmark &  The number of attacks during treatment periods were \hl{31} and \hl{23}\\ 
    \textbf{Qualifier} & \xmark & \xmark & Pindolol and metoprolol lowered blood pressure \hl{to the same extent}\\ 
    \bottomrule
    \end{tabular}
    \caption{Examples of entity types in our schema. EBM and CTKG columns indicate whether these entity types are present in the EBM-NLP and CTKG schemas respectively. EBM-NLP uses IE to extract information according to its schema, while CTKG is a database schema not based on IE.}
    \label{tab:entity_types}
\end{table*}

Some prior efforts have also explored result extraction from biomedical literature. The EBM-NLP \cite{nye2018corpus} and Evidence Inference \cite{lehman2019inferring} corpora contain annotations for experimental findings from clinical trials, following the well-established PICO (participant, intervention, comparator, outcome) framework \cite{richardson1995well}. \citet{sanchez2022annotated} also develop a PICO-inspired schema-based annotation format for diabetes and glaucoma trials. \citet{chen2022knowledge} focuses on aggregating findings, which are already manually organized in structured format in databases such as AACT (Aggregate Analysis of ClinicalTrials.gov) \citep{tasneem2012database}. However, these efforts are tailored to clinical trials and do not translate easily to other domains. Finally, \citet{luo2022biored} conducted \emph{novelty} annotation for relations, indicating whether they were presented as new observations; however they did not focus on experimental findings.

In contrast, we develop a representation schema expressive enough to capture fine-grained experimental findings, while generalizing across scientific domains. Our schema also contains phenomena challenging for SOTA IE models (\S\ref{ssec:complex}).

\subsection{Extracting Numeric Information}
Another unique aspect of our schema is our focus on capturing numeric information from experimental findings and setup, which is understudied. Some prior work on open IE has explored extraction and linking of numeric spans \cite{madaan2016numerical,saha-etal-2017-bootstrapping}, including linking to implied entities \cite{elazar2019s} (e.g., ``it's worth two million'' can be linked to currency). However, these models broadly focused on sentence-level extraction and did not evaluate on scientific text.

Within the scientific domain, some studies have focused on numeric information extraction from biomedical/clinical text. \citet{kang2013extracting} and \citet{claveau2017numerical} extract numeric spans from FDA-released descision summaries and clinical trial eligibility criteria respectively. EBM-NLP \cite{nye2018corpus} annotates some categories of numeric information associated with cohorts participating in a clinical trial, but ignores trial outcomes and findings. Among non-medical scientific domains, numeric span extraction work has mainly focused on extraction from tables \cite{hou-etal-2019-identification}. None of these studies focus extensively on linking numeric spans with entities that can help in interpreting this information, which is key to our work.

\section{Annotation Schema}

\begin{table}[]
\setlength{\tabcolsep}{2.25pt}
    \centering
    \small
    \begin{tabular}{lccp{2.5cm}}
    \toprule
    \textbf{Type} & \textbf{EBM} & \textbf{CTKG} & \textbf{Example} \\ \midrule
     Age & \cmark & \xmark & for those \hl{age 60-67 years} \\
     Sex & \cmark & \xmark & \hl{210} females\\
     Size & \cmark & \cmark & \hl{12} patients\\
     Condition & \cmark & \cmark & patients getting \hl{hemodialysis}\\
     Demographic & \xmark & \xmark & A 40's \hl{Japanese} man \\ \midrule
     Route & \xmark & \xmark & \hl{oral} verapamil \\
     Dosage & \xmark & \xmark & verapamil \hl{480 mg/day} \\
     Strength & \xmark & \xmark & rizatriptan \hl{5 mg}\\
     Duration & \xmark & \xmark & for \hl{4 weeks} \\
    \bottomrule
    \end{tabular}
    \caption{Examples of attribute types in our schema. EBM and CTKG columns indicate whether these types are present in the EBM-NLP and CTKG schemas.}
    \label{tab:attribute_types}
    \vspace{-2em}
\end{table}

\begin{table*}[]
\setlength{\tabcolsep}{2.25pt}
    \centering
    \small
    \begin{tabular}{llccp{10cm}}
    \toprule
    \textbf{Type} & \textbf{Arity} & \textbf{EI} & \textbf{CTKG} & \textbf{Example} \\ \midrule
    AttributeOf & N-ary & \xmark & \xmark & ($Subpopulation$: 144 had the U-type method, $Size$: 144) \\
    SubpopulationOf & N-ary & \xmark & \xmark & ($Population$: 285 women, $Subpopulation$: 144 had the U-type method, $Subpopulation$: 141 had the H-type method)\\
    TreatmentOf & Binary & \xmark & \cmark & ($Subpopulation$: 144 had the U-type method, $Treatment$: U-type method)\\
    Result & N-ary & \cmark & \cmark & ($Subpopulation$: 144 had the U-type method, $Measurement$: objective cure rates, $NumericFinding$: 87.5\%)\\
    \bottomrule
    \end{tabular}
    \caption{Examples of relation types in our schema. EI and CTKG columns indicate whether these relation types are present in the EI and CTKG schemas respectively. While the EI and CTKG datasets contain 4-ary and binary result relations respectively, our n-ary schema allows fine-grained information to be captured more flexibly.}
    \label{tab:relation_types}
\end{table*}

We develop a new annotation schema to represent fine-grained clinical findings present in biomedical abstracts, and later demonstrate its broader applicability to domains beyond biomedicine (\S\ref{ssec:schemastudy}). Our schema captures this knowledge via three main elements, commonly used in IE tasks:  

\noindent
\textbf{1. Entities} involved in a study, which are spans of text, either contiguous or non-contiguous, belonging to one of the seven types listed in Table~\ref{tab:entity_types}.

\noindent
\textbf{2. Attributes} associated with entities, which are also contiguous or non-contiguous spans of text, belonging to one of the nine types listed in Table~\ref{tab:attribute_types}. The first five attribute types are associated with population and subpopulation entities, while the remaining four types are associated with treatment entities. Other entity types do not have any associated attributes. 

\noindent
\textbf{3. N-ary Relations} linking together various entities and attributes, where N (relation arity) is variable and nesting is allowed. A relation is an n-tuple, where each element can be an entity, attribute or another n-ary relation. Relations are categorized into four types listed in Table~\ref{tab:relation_types}.


\subsection{Comparison to Existing Clinical Schemas}
Prior work such as EBM-NLP \citep{nye2018corpus} and Evidence Inference \citep{lehman2019inferring,deyoung2020evidence} has focused on developing IE schemas to represent clinical knowledge appearing in the literature in a structured format. In addition, work such as CTKG \citep{chen2022knowledge} outside the NLP/IE sphere has built schema for representing clinical information in databases. However, these schemas suffer from a few shortcomings: (i) most are designed for clinical trials; their applicability to other types of biomedical literature is untested, (ii) focus on a small set of broad entity types, which leaves out fine-grained details, (iii) follow strict relation formats, which makes it hard to capture additional nuance that might be useful for interpreting findings. 


Our schema makes several enhancements to tackle these issues. First, it is extensible to other categories of biomedical literature beyond clinical trials, and we demonstrate this by applying our schema to case reports. Second, our schema captures more fine-grained information about various entities than prior work via attributes (see Table~\ref{tab:attribute_types}). Third, allowing for variable arity and nesting in relation annotation provides the flexibility which makes our schema capable of representing both atomic findings (e.g., value of primary outcome observed for a given treatment) as well as composite findings (e.g., outcome improvement observed for treatment vs control groups). Tables~\ref{tab:entity_types},~\ref{tab:attribute_types} and~\ref{tab:relation_types} provide a more detailed comparison of our schema with EBM-NLP, EI and CTKG.



\subsection{Annotation Complexity}
\label{ssec:complex}
In addition to using an expanded set of entity, attribute and relation types, our annotation schema supports the following phenomena (also illustrated in Figure~\ref{fig:trialanno}), unifying them all in a single dataset:\\
\textbf{Discontinuous spans:} Biomedical abstracts often present multiple entities as conjunctive phrases or lists of items, so we allow discontinuous span annotation to capture every entity. For example, given the phrase ``maximal diameters and volumes'', our scheme captures two measurement entities: ``maximal diameters'' and ``maximal volumes'', with the latter being a discontinuous span.\\
\textbf{Nested/overlapping spans:} Attributes, as defined in our annotation scheme, are often present within an entity span or overlap with an entity span. This motivates our decision to allow nested and overlapping spans to be annotated.\\ 
\textbf{Variable arity in relations:} Owing to variation in clinical studies, findings are often described in a wide range of formats (e.g., outcome for a single population, outcome for a pair of populations, outcome for a single population at different time periods, etc.). This diversity motivated our choice of \emph{variable arity} for relation annotation, similar to \citet{tiktinsky2022dataset}.\\
\textbf{Nested relations:} In addition to outcomes for individual populations/groups, clinical studies often present comparative findings and analyses, such as improvement on an outcome given a pair of interventions. Our scheme allows for annotation of nested relations to link these higher-order observations with their associated atomic findings.\\
Our complete annotation guidelines can be found at \url{https://github.com/allenai/CARE}. Figure~\ref{fig:trialanno} presents partial entity, attribute and relation annotations for an example clinical trial abstract.

\section{Dataset Collection}


\noindent
\textbf{Annotation Tool:} We use TeamTat\footnote{\url{https://www.teamtat.org}} \cite{islamaj2020teamtat}, a web-based tool for team annotation since it allows for n-ary and nested relation annotation, a core component of our schema.

\noindent
\textbf{Annotator Background:} We recruit two in-house annotators\footnote{included as co-authors on this paper} with backgrounds in data analytics and data science, both having extensive experience in reading and annotating scientific papers. One of our annotators has a background in biology. Both annotators went through several pilot rounds to gain familiarity with our task and schema. Additionally, we used their feedback and insights from pilots to solidify our schema design (see~\S\ref{ssec:pilots}). We also solicited feedback from two medical students and an MD to validate our final schema. 

\noindent
\textbf{Data Sources:} \dataset{} covers two categories of biomedical literature: (i) clinical trials, and (ii) case reports. Clinical trials are research studies that test a medical, surgical, or behavioral intervention in people to determine whether a new form of treatment or prevention or a new diagnostic device is effective. Case reports are detailed reports of the symptoms, signs, diagnosis, treatment, and follow-up of an individual patient, usually motivated by unusual or novel occurrences. We sample clinical trials from the EBM-NLP \cite{nye2018corpus} dataset, which consists of 4993 abstracts annotated with PICO spans, only retaining abstracts containing at least one number (4685 in total). To sample case reports, we extract all reports with at least one number in the abstract from PubMed (907,862 in total) and randomly sample from this pool. We sample 350 abstracts from each source, resulting in our final dataset size of 700 abstracts, which is slightly larger than other prior corpora that perform fine-grained annotation (\S~\ref{ssec:datastatssec}). Further characteristics of our abstract sample are detailed in Appendix~\ref{sec:dataprocess}.


\begin{table}[]
    \centering
    \small
    \begin{tabular}{lcc}
     \toprule  \textbf{Category} & \textbf{Exact F1} & \textbf{Partial F1}\\ \midrule
       Entity  & 0.5764 & 0.7578 \\
       Attribute  & 0.6174 & 0.7801 \\
       Relation  & 0.4209 & 0.7414 \\ \bottomrule
    \end{tabular}
    \caption{Final inter-annotator agreement scores on a sample of 28 abstracts, measured during full-scale data annotation. }
    \label{tab:finalagreement}
    \vspace{-1.5em}
\end{table}

\subsection{Annotation Pilots}
\label{ssec:pilots}
We conducted three pilot rounds with the following goals: (i) training annotators to apply our schema, (ii) evaluating agreement, and (iii) assessing whether our schema captures clinical knowledge of interest. Annotators worked on a fresh set of 5-10 abstracts per round, followed by agreement computation and disagreement discussion. For entity and attribute annotation, agreement is computed as entity-level F1 between annotators, using both strict (entity boundaries match exactly) and partial (entity boundaries overlap on at least one token) matching. For relations, we first align annotations from both annotators by linking pairs of relations which share $\geq50\%$ of participating entities. Agreement is computed as F1 score between annotators, using both strict ($100\%$ of entities match) and partial matching. After achieving reasonable agreement levels by round 3 (partial F1 scores of 0.79, 0.68 and 0.79 for entity, attribute and relation annotation respectively), we started full-scale data annotation (further discussion in Appendix~\ref{sec:dataprocess}).

\begin{table}[]
    \centering
    \small
    \begin{tabular}{lccc}
    \toprule \textbf{Metric}  & \textbf{Train} & \textbf{Dev} & \textbf{Test}\\ \midrule
    \textbf{\#Docs} & 500 & 100 & 100\\ 
    \textbf{\#Tokens}  & 135,363 & 27,120 & 25,219 \\ 
    \textbf{\#Entities}  & 12022 & 2367 & 2286 \\ 
    \textbf{\#Attributes}  & 3992 & 804 & 762 \\ 
    \textbf{\#Relations}  & 8205 & 1594 & 1560 \\ \bottomrule
    \end{tabular}
    \caption{Statistics for final collected dataset.}
    \label{tab:datastats}
\end{table}
\subsection{Full-Scale Annotation}
The full-scale data annotation process was conducted in six rounds. To continue monitoring agreement, a small agreement set of 5 abstracts (not identified to the annotators) was included in every round. Table~\ref{tab:roundagreement} in the appendix presents inter-annotator agreement during each annotation round, while Table~\ref{tab:finalagreement} shows overall agreement scores. Overall and per-round agreement scores continued to remain in the same range as agreement scores from later pilot rounds, demonstrating consistency in annotation quality. Despite the complexity of our schema, our agreement scores are comparable to datasets using simpler schemas like EBM-NLP (entity agreement of 0.62-0.71; Cohen's kappa) and SciERC (relation agreement of 67.8; kappa score). Appendix~\ref{sec:dataprocess} provides additional details about our full-scale annotation setup.

\noindent
\textbf{Consensus Annotation: }For all abstracts annotated by multiple annotators during pilots or full-scale annotation (55 in total), we construct a ``consensus'' version post disagreement discussion. The final dataset releases consensus annotations for these abstracts. Since this subset has been annotated by multiple annotators and discussed extensively, we expect annotations to be higher-quality and include all these abstracts in the test set.  

\begin{table}[]
    \centering
    \small
    \begin{tabular}{lccc}
    \toprule \textbf{Phenomenon}  & \textbf{Train} & \textbf{Dev} & \textbf{Test}\\ \midrule
    \textbf{\#Discontinuous Spans} & 8.9\% & 10.1\% & 9.3\%\\ 
    \textbf{\#Nested Spans}  & 3.4\% & 4.3\% & 2.5\%\\ 
    \textbf{\#Overlapping Spans}  & 1.6\% & 2.0\% & 0.7\%\\ 
    \textbf{\#Nested Relations}  & 11.4\% & 11.2\% & 11.9\%\\ \bottomrule
    \end{tabular}
    \caption{Prevalence of interesting annotation phenomena in final collected dataset.}
    \label{tab:phenomstats}
    \vspace{-1.5em}
\end{table}


\subsection{Dataset Statistics}
\label{ssec:datastatssec}
Table~\ref{tab:datastats} gives an overview of statistics for our final collected dataset. Our dataset size is comparable to other prior biomedical corpora which performs exhaustive fine-grained annotation (though not always with a clinical knowledge focus) such as BioRED (\citet{luo2022biored}; 600 abstracts) and \citet{sanchez2022annotated} (211 abstracts). Table~\ref{tab:phenomstats} presents the proportion of various interesting phenomena allowed by our schema in the final dataset. Interestingly, \dataset{} contains ~9\% discontinuous spans, making it one of the rare datasets containing a large proportion of discontinuous mentions.\footnote{\citet{dai-etal-2020-effective} considers 10\% discontinuous spans to be a high proportion, identifying only three biomedical datasets that satisfy this criterion: CADEC \citep{karimi2015cadec}, ShARe 13 \citep{pradhan2013task} and ShARe 14 \citep{mowery2014task}.} At ~11\%, the final data also contains a high proportion of nested relations. 

\section{Benchmarking IE Models}
We benchmark the performance of two categories of models on \dataset{}:
(i) extractive models, and (ii) generative LLMs. We also test generative LLMs in two settings: (i) finetuning on the full training set, and (ii) zero-shot and in-context learning.

\noindent
\textbf{Experimental Setup:} We test each model on the three sub-tasks---entity extraction, attribute extraction and relation extraction---in isolation. Model performance on entity and attribute extraction is evaluated using entity-level F1. Relation extraction performance is evaluated using a relaxed overlap F1 score metric inspired by \citet{tiktinsky2022dataset}, which assigns partial credit to correctly identified subsets of entities in a relation, even if all identified entities do not match. As with agreement score calculation, predicted relations are first aligned with gold relations by choosing the gold relation with highest overlap per predicted relation. Then a partial match score is computed as $\#shared\_entities/total\_entities$ and used in the F1 computation instead of binary 0/1 score. 


\subsection{Extractive IE Baselines:}
We evaluate the following systems:
\begin{itemize}[leftmargin=*,topsep=0pt]
\setlength\itemsep{-0.5em} 
    \item \textbf{OneIE} \cite{lin-etal-2020-joint}: A sentence-level joint entity, relation and event extraction system, which extracts an ``information network'' representation of entities and events (nodes), connected by relations (edges). Beam search is used to find the highest-scoring network.
    \item \textbf{PURE} \cite{zhong-chen-2021-frustratingly}: A sentence-level pipelined extraction system, which learns separate contextual representations for entity and relation extraction, using entity representations to further refine relation extraction.
    \item \textbf{LocLabel} \cite{shen-etal-2021-locate}: A sentence-level two-stage named entity recognition (NER) system capable of extracting nested spans. Inspired by object detection work, it produces boundary proposals for candidate entities, then labels them with correct entity types.
    \item \textbf{W2NER} \cite{li2022unified}: A sentence-level unified NER model, capable of extracting nested and discontinuous spans. It recasts NER as word-word relation classification on a 2-D grid of word pairs, then decodes word pair relations into final span extractions.
\end{itemize}

For comparability and better adaptation to our dataset, we replace BERT-based encoders in all systems with PubmedBERT \cite{gu2021domain}, and follow best-reported hyperparameters per system (see Appendix~\ref{sec:hyperparam}). Table~\ref{tab:results} presents their performance on entity and attribute extraction. Unfortunately, applying these systems to our relation extraction task is infeasible, since none of them are designed for document-level relation extraction or n-ary relations. \citet{tiktinsky2022dataset} modify PURE for n-ary relation extraction with variable arity. However, given a set of candidate entities, they consider all possible n-ary combinations and predict relationships per cluster. This is tractable for their work on sentence-level extraction of single-type (drug interaction) relations, but not tractable for document-level multi-type n-ary relation extraction.\footnote{On limiting combination size to 10, every abstract produces ~500,000 candidate combinations} Therefore, we do not test extractive models on relation extraction. 

\begin{table}[]
    \centering
    \small
    \begin{tabular}{lccc}
    \toprule \textbf{Model} & \textbf{Ent F1} & \textbf{Attr F1} & \textbf{Rel F1} \\ \midrule
    \multicolumn{4}{c}{\textbf{Extractive Baselines}} \\ \midrule
    \textbf{OneIE} & 55.07 & 48.84 & -- \\
    \textbf{PURE} & \textbf{55.94} & \textbf{61.04} & -- \\
    \textbf{LocLabel} & 53.69 & 55.25 & -- \\
    \textbf{W2NER} & 51.84 & 57.98 & -- \\
    \midrule \multicolumn{4}{c}{\textbf{Generative Baselines}} \\ \midrule
    \textbf{FLAN-T5} & \textbf{45.08} & 23.27 & 33.24 \\
    \textbf{BioGPT} & 14.43 & \textbf{29.84} & 33.15 \\
    \textbf{BioMedLM} & 1.50 & 10.62 & 32.76 \\ \midrule
    \textbf{GPT-3.5 0-shot} & 11.14 & 5.06 & 14.35 \\
    \textbf{GPT-3.5 1-shot} & 21.40 & 8.61 & 31.58 \\
    \textbf{GPT-3.5 3-shot} & 23.40 & 8.85 & 31.58 \\
    \textbf{GPT-3.5 5-shot} & 8.92 & 9.92 & 32.20 \\ \midrule
    \textbf{GPT-4 0-shot} & 26.89 & 9.02 & 32.04 \\
    \textbf{GPT-4 1-shot} & 31.07 & 11.82 & 42.81 \\
    \textbf{GPT-4 3-shot} & 16.68 & 13.16 & 53.69 \\
    \textbf{GPT-4 5-shot} & 5.04 & 13.90 & \textbf{55.04} \\
    \bottomrule
    \end{tabular}
    \caption{Performance of all extractive and generative baselines on entity, attribute and relation extraction.}
    \label{tab:results}
    \vspace{-1.5em}
\end{table}


Another caveat with extractive models is that they do not identify discontinuous spans (except W2NER). To assess how this impacts model performance, we compute an additional entity-level F1 score which merges span predictions linked in gold annotation (i.e., we assume oracle span merging), and observe that this does not significantly improve performance (avg. increase of $\sim$1.5 F1). Therefore, Table~\ref{tab:results} reports F1 scores without merging.


\subsection{Generative IE Baselines:}
Motivated by recent work demonstrating LLM capabilities on information extraction \cite{wadhwa-etal-2023-revisiting}, we assess the ability of LLMs on our tasks, in both finetuning and zero-shot/in-context learning settings.


We evaluate the following finetuned LLMs:
\begin{itemize}[leftmargin=*,topsep=0pt]
\setlength\itemsep{-0.5em} 
    \item \textbf{FLAN-T5} \cite{chung2022scaling}: Enhanced version of T5 \cite{raffel2020exploring} finetuned on a large mixture of tasks, but not specifically pretrained for biomedicine. We use FLAN-T5-XL, which has 3B parameters.
    \item \textbf{BioGPT} \cite{luo2022biogpt}: A 1.6B autoregressive model, pretrained from scratch on 15M abstracts and titles from PubMed with a custom Pubmed-trained tokenizer.
    \item \textbf{BioMedLM}\footnote{\url{https://crfm.stanford.edu/2022/12/15/biomedlm.html}}: A 2.7B autoregressive model, pretrained from scratch on all PubMed abstracts and full-texts from the Pile \cite{gao2020pile} with a custom PubMed-trained tokenizer.
\end{itemize}

\begin{table*}[]
    \centering
    \small
    \begin{tabular}{p{2.3cm}p{3cm}p{9.5cm}}
    \toprule \textbf{Original Type} & \textbf{Generalized Type} & \textbf{Description} \\ \midrule
    Population & Research Problem Context & Setting/scenario in which the authors are testing their hypothesis (e.g., task or dataset being studied in ML/NLP). \\
    Subpopulation & Problem Stages/Sub-parts & Subgroups or subsamples of overall setting (e.g., dataset splits in ML/NLP). \\
    Treatment & Technique/Method & Key technique being proposed or investigated and other techniques being compared (e.g., model or metric in ML/NLP). \\ \midrule
    SubpopulationOf & Sub-PartOf & Links together problem context entities to stage/sub-part entities (e.g., for ML/NLP, this relation would link the overall task to low-data and fully supervised settings). \\
    TreatmentOf & AppliedTo & Links together a technique to all the problem contexts/sub-parts it is being tested in. \\
    \bottomrule
    \end{tabular}
    \caption{Changes required to construct a generalized version of our original schema developed for clinical finding extraction, which we use to test whether it applies to other domains such as computer science and materials science}
    \label{tab:genschema}
    \vspace{-1.5em}
\end{table*}

When training and testing on attribute and relation extraction, these models are provided gold entities and attributes by surrounding them with entity markers ($<ent> </ent>$) in the input. 


We evaluate GPT3.5 and GPT4 in zero-shot and in-context learning settings. We provide our IE schema and example outputs and prompt the model to produce extractions in a clean JSON format that adheres to the schema. Additionally, for our in-context learning experiments, we follow \cite{liu2021makes} and select the $k$ \emph{most similar} examples from the training set for every test instance according to similarity computed by the SPECTER v2.0 \cite{singh2022scirepeval} PRX model trained on scientific titles and abstracts. Selected examples are appended to the prompt in decreasing order of similarity, with later examples dropped if they don't fit. We run experiments for the $k=1, 3, 5$ most similar examples. Further hyperparameter details for all models are provided in Appendix~\ref{sec:hyperparam} and full prompts are provided in Appendix~\ref{sec:prompts}.

Table~\ref{tab:results} shows the performance of all generative models. One caveat with GPT3.5/4 is that model outputs sometimes contain correct entity/attribute spans assigned to the wrong type (e.g., a subpopulation misclassified as a population entity in a result relation). Since we are evaluating the performance of relation extraction in isolation, we do not consider such mistyping as errors. 

\subsection{End-to-End Evaluation:}
In addition to evaluating SOTA systems on each sub-task in isolation, we assess the feasibility of end-to-end extraction. Table~\ref{tab:results} shows that PURE is the best-performing system on entity and attribute extraction. On the other hand, GPT4 5-shot and FLAN-T5 perform best on relation extraction (GPT3.5 5-shot and BioGPT are close). We test out a hybrid end-to-end extraction system in which entities and attributes are detected using PURE, then input text marked up with these extractions is provided to FLAN-T5 for relation extraction. This hybrid system achieves an F1 score of 33.58, very similar to RE performance with gold markup. Hypothesizing that this might be an indication that finetuned LLMs ignore entity/attribute markup during RE, we run an additional experiment in which we train FLAN-T5 to extract relations from raw text (no markup). This setup achieves an F1 score of 33.07, showing that entity/attribute markup does not provide significant benefit.



\section{Discussion}
\subsection{How much does strict evaluation underestimate LLM performance?}
Table~\ref{tab:results} shows that even fully-supervised generative models severely lag behind much smaller extractive models on entity and attribute extraction. However, prior work \cite{wadhwa-etal-2023-revisiting} has observed that strict IE evaluation metrics underestimate the performance of LLMs since their outputs often contain minor variations from gold annotations, which could still be correct. Therefore, we conduct human evaluation of a subset of FLAN-T5 and GPT4 5-shot predictions on entity and attribute extraction for a more accurate assessment.

For every setting, we collect all abstracts with one or more wrong predictions and randomly sample ten to evaluate. We go over all false positives per abstract marking ones that could be considered correct. Our evaluation shows that for FLAN-T5, 35 out of 73 entity and 12 out of 32 attribute errors are marked correct. For GPT4, these numbers are worse; 38 out of 126 entity and 20 out of 79 attribute errors are marked correct. This indicates that LLMs indeed struggle with our span extraction tasks, and their poor performance is not simply a consequence of strict evaluation.

\subsection{How easily can we extend our schema to other domains?}
\label{ssec:schemastudy}

\begin{figure}
\centering
\includegraphics[scale=0.5]{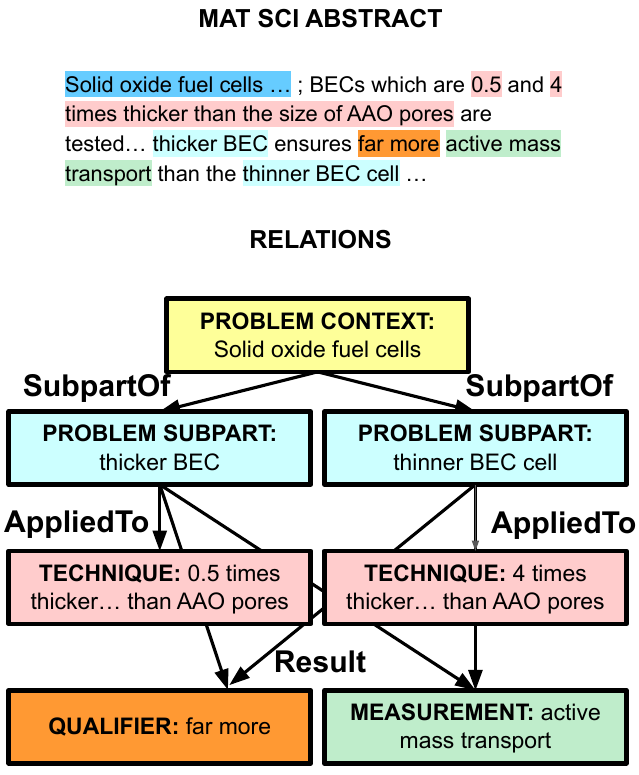}
\caption{A partial example of entity, attribute and relation annotation using our generalized schema for a materials science abstract.} 
\label{fig:matscianno}
\vspace{-1.5em}
\end{figure}
Though we focus on extracting clinical findings from biomedical literature during schema design, we try to incorporate enough flexibility to allow our schema to be easily adapted to other scientific domains. To demonstrate this flexibilty, we conduct small-scale pilots in two additional domains: (i) Computer Science, and (ii) Materials Science. 

We first develop a \emph{generalized} version of our proposed schema for these studies. Of the three elements in our schema, entities and relations are largely transferable and only require minor renaming. Table~\ref{tab:genschema} provides an overview of changes made to entity/relation nomenclature. Attributes on the other hand, were tailored more closely to our goal of extracting clinical findings. Therefore, we drop all attributes and ask our annotators to propose candidate attributes as they go through the annotation process. We use the same annotators who participated in dataset creation, to leverage their existing familiarity with our schema, assigning one annotator to each domain. Their task is to annotate ten abstracts each while documenting: (i) potential attributes that can be added to the schema, and (ii) important experimental information missed by the generalized schema. 

After completing the task, annotators reported that it was feasible to apply our proposed schemas to these scientific domains. Figure~\ref{fig:matscianno} shows an example materials science abstract with partial annotations according to the generalized schema. Computer science posed some difficulty due to the presence of lots of relative results and references in the abstract, which made entity annotation ambiguous. However, there were no important aspects of experimental information, aside from potential attribute proposals, that our current schema could not account for. 

\section{Conclusion}
In this work, we presented \dataset{}, a new IE dataset for the task of extracting clinical findings from biomedical literature. To collect this dataset, we first developed a new annotation schema capable of capturing fine-grained information about experimental findings, which unified several challenging IE phenomena such as discontinuous spans, nested relations and variable arity n-ary relations. Using this annotation scheme, we collected an extensively annotated dataset of 700 abstracts from clinical trials and case reports. Our benchmarking experiments showed that state-of-the-art extractive and generative LLMs including GPT4 still struggle on this task, particularly on relation extraction. We release both our annotation schema and \dataset{} as a challenging new resource for the IE community and to encourage further research on extraction and representation of findings from scientific literature.

\section{Limitations}
Despite being a cornerstone of our work, the richness and complexity of our newly proposed annotation schema also poses some limitations. Annotators needed some prior experience with reading and understanding complex scientific text, and had to undergo multiple rounds of additional training before they were able to accurately apply our schema and start full-scale annotation. Though these stringent expertise and training requirements and heavy reliance on human annotators helped us collect a high-quality resource in \dataset{}, they simultaneously limit the scalability of our collection protocol and make it difficult to construct large-scale benchmarks for this task, spanning multiple domains/fields of science. 

Our annotated corpus, \dataset{}, is based on RCTs and case reports. While our schema is broad and expressive enough to generalize to other experimental domains with minor adaptations, our generalization annotation studies were comparatively small and preliminary, limited to testing the schema on computer science and material science papers. In addition, while our schema covers many types of experimental findings, the richness and huge variety of scientific experiments neccessarily means that more types of findings could be added. In the future, more studies should be performed on using our schema in other domains, and on extending our schema with more types of information (entities, attributes, relations). \dataset{} also focuses on English-language papers only, and in the future it would be interesting and important to extend our schema and dataset to cover biomedical/clinical studies in other languages, to capture important scientific findings that are potentially missed when only looking at papers in English.

Finally, a limitation of our current benchmarking effort is the lack of more flexible evaluation metrics, particularly when assessing the performance of generative LLMs. We try to provide supplementary human evaluation for some models to overcome this issue, but this is not scalable and would require ongoing/continuous evaluation efforts. This is not a major focus for our current work, but developing more flexible automated evaluation is an important future direction for IE research.


\section*{Acknowledgements}
This work was funded in part by NSF Convergence Accelerator Award ITE-2132318. We would like to thank Lucy Lu Wang, Byron Wallace and Dr. Ruth Sapir-Pichhadze for valuable feedback on early versions of the annotation schema, and the Semantic Scholar team at AI2 and our anonymous reviewers for their helpful comments on our work.

\bibliography{custom}
\bibliographystyle{acl_natbib}

\appendix
\section{Schema Definitions}
\subsection{Entity Types}
Entities can belong to one of the following seven types:
\begin{enumerate}[leftmargin=*,topsep=0pt]
\setlength\itemsep{-0.5em}    
    \item \textbf{Population:} Patient groups/cohorts studied in an article.
    \item \textbf{Subpopulation:} Slices/sub-groups of a population entity sharing some underlying characteristic.
    \item \textbf{Treatment:} Treatment regimens, procedures, therapies etc. prescribed and/or tested to alleviate a population's conditions/symptoms.
    \item \textbf{Measurement:} Tests used to assess population status and outcomes of the tested intervention.
    \item \textbf{Temporal:} Temporal information such as time points at which outcomes are measured. 
    \item \textbf{NumericFinding:} All numeric information associated with study findings (e.g., p-values, hazard ratios, etc.). 
    \item \textbf{Qualifier:} Non-numeric information associated with study findings that provides important perspective for interpreting them (e.g., phrases indicating evidence directionality).
\end{enumerate}

\subsection{Attribute Types}
Attributes can belong to one of the following nine types:
\begin{enumerate}[leftmargin=*,topsep=0pt]
\setlength\itemsep{-0.5em}    
    \item \textbf{Age:} Numeric or non-numeric information about the age of the population under study.
    \item \textbf{Sex:} Reported sex of the population under study.
    \item \textbf{Size:} Size of the population sample under study.
    \item \textbf{Condition:} Medical conditions prevalent in the study population, including diseases, symptoms, prior medical history and procedures, etc.
    \item \textbf{Demographic:} Additional demographic information reported about the population such as location, race, etc.
    \item \textbf{Route:} Description of the way an intervention is administered (e.g., a chemical may be administered orally, topically, intravenously, etc.).
    \item \textbf{Dosage:} Quantity of administration for the intervention being studied. This is not necessarily limited to chemical/drug interventions (e.g., for an intervention like educational sessions, number of sessions is considered ``dosage''). 
    \item \textbf{Strength:} Strength of chemical/drug interventions administered.
    \item \textbf{Duration:} Interval of time over which an intervention was administered.
\end{enumerate}

\subsection{Relation Types}
Our schema allows for both binary and n-ary relations (with variable n), to capture four types of structure:
\begin{enumerate}[leftmargin=*,topsep=0pt]
\setlength\itemsep{-0.5em} 
    \item \textbf{AttributeOf:} N-ary relations linking population and intervention entities with their associated attributes.
    \item \textbf{Subpopulation:} N-ary relations capturing parent-child relationships between population and subpopulation entities.
    \item \textbf{InterventionOf:} Binary relations linking population and subpopulations entities with the intervention(s) tested on them.
    \item \textbf{Result:} N-ary relations capturing all numeric or non-numeric outcome results and comparisons reported by linking together the population, subpopulation, intervention, measurement, numericfinding and/or qualifier and temporal entities involved in each result/comparison.
\end{enumerate}
All n-ary relations can contain multiple entities of a single type. For example, a result relation can involve multiple interventions or populations. The only cardinality constraints imposed are that every result relation should focus on a \emph{single} measurement entity and always contain \emph{at least one} population/intervention entity. 

\section{Additional Annotation Rules}
While using this annotation schema to annotate clinical knowledge, we also keep in mind the following rules:
\begin{itemize}[leftmargin=*,topsep=0pt]
\setlength\itemsep{-0.5em}
\item For every entity/attribute span, only annotate its first occurrence in the text, unless there is a more descriptive span later. We follow this rule to avoid conducting an additional coreference annotation step to link all spans referring to the same entity.
\item Ignore misspellings and include all associated modifiers and abbreviations while annotating spans
\item Do not annotate generic or high-level spans (e.g., genetic disorder), or generic terms (e.g., complications, deficiency, disease, syndrome, gene, drug, protein, nucleotide, etc.).
\item Do not annotate background occurrences of entities. For example, if a treatment Y is mentioned as “X is usually treated using Y,...”, do not annotate Y unless Y was one of the treatments actually given to a population in the current study.
\end{itemize}

\begin{table*}[t]
    \centering
    \begin{tabular}{ccccccc}
     \toprule  \textbf{Round} & \multicolumn{2}{c}{\textbf{Entity F1}} & \multicolumn{2}{c}{\textbf{Attribute F1}} & \multicolumn{2}{c}{\textbf{Relation F1}} \\ \cmidrule{2-7}
     & \textbf{Exact} & \textbf{Partial} & \textbf{Exact} & \textbf{Partial} & \textbf{Exact} & \textbf{Partial} \\ \midrule
     Pilot 1 & 0.6240 & 0.7579 & 0.7215 & 0.8163 & 0.2193 & 0.6379 \\
     Pilot 2 & 0.7206 & 0.8818 & 0.6923 & 0.7385 & 0.4997 & 0.7878 \\
     Pilot 3 & 0.6449 & 0.7900 & 0.5370 & 0.6852 & 0.4449 & 0.7960 \\ \midrule
     Batch 1 & 0.5130 & 0.7318 & 0.7611 & 0.8496 & 0.3899 & 0.6979 \\
     Batch 2 & 0.6094 & 0.7900 & 0.6216 & 0.8508 & 0.6397 & 0.9137 \\
     Batch 3 & 0.5312 & 0.7797 & 0.6364 & 0.8182 & 0.3121 & 0.7595 \\
     Batch 4 & 0.5714 & 0.7817 & 0.7347 & 0.7755 & 0.5399 & 0.7343 \\
     Batch 5 & 0.5643 & 0.6929 & 0.4717 & 0.6762 & 0.3382 & 0.6766 \\
     Batch 6 & 0.6358 & 0.7930 & 0.5417 & 0.7582 & 0.3122 & 0.6890 \\ \midrule
     Overall & \textbf{0.5764} & \textbf{0.7578} & \textbf{0.6174} & \textbf{0.7801} & \textbf{0.4209} & \textbf{0.7414} \\ \bottomrule
    \end{tabular}
    \caption{Evolution of inter-annotator agreement during pilots and full-scale annotation rounds}
    \label{tab:roundagreement}
\end{table*}

\begin{table}[]
    \centering
    \begin{tabular}{ccc}
    \toprule \textbf{Type} & \textbf{Exact F1} & \textbf{Partial F1}\\ \midrule
    Population  & 0.4333 & 0.8665 \\
    Subpopulation  & 0.4299 & 0.6168 \\
    Intervention  & 0.4333 & 0.5781 \\
    Measurement  & 0.5230 & 0.7554 \\
    Temporal  & 0.6230 & 0.6885 \\
    NumericFinding & 0.7063 & 0.8812 \\
    Qualifier  & 0.6911 & 0.7749 \\ \bottomrule
    \end{tabular}
    \caption{Inter-annotator agreement per entity type}
    \label{tab:entityagreement}
\end{table}

\begin{table}[]
    \centering
    \begin{tabular}{ccc}
    \toprule \textbf{Type} & \textbf{Exact F1} & \textbf{Partial F1}\\ \midrule
    Age  & 0.8500 & 0.9756 \\
    Sex  & 0.9231 & 0.9231 \\
    Size & 0.6462 & 0.7385 \\
    Condition  & 0.5091 & 0.7429 \\
    Demographic  & 0.6667 & 0.8000 \\
    Route & 0.8000 & 0.8000 \\
    Dosage  & 0.6923 & 0.9630 \\ 
    Strength & - & - \\
    Duration & 0.0800 & 0.4800 \\ \bottomrule
    \end{tabular}
    \caption{Inter-annotator agreement per attribute type. Note that the agreement sample did not include any strength entities.}
    \label{tab:attributeagreement}
\end{table}

\begin{table}[]
    \centering
    \begin{tabular}{ccc}
    \toprule \textbf{Type} & \textbf{Exact F1} & \textbf{Partial F1}\\ \midrule
    AttributeOf & 0.7654 & 0.7654 \\
    InterventionOf & 0.3797 & 0.3797 \\
    SubpopulationOf & 0.1633 & 0.5185 \\
    Result & 0.2561 & 0.7994 \\ \bottomrule
    \end{tabular}
    \caption{Inter-annotator agreement per relation type}
    \label{tab:relationagreement}
\end{table}

\section{Dataset Construction Details}
\label{sec:dataprocess}
\noindent
\textbf{Characteristics of sampled abstracts}: Since the EBM-NLP corpus sampled randomized clinical trials from PubMed with an emphasis on cardiovascular diseases, cancer, and autism, the clinical trials portion of our dataset also heavily features these topics. On the other hand, for case reports, comparing MeSH term distributions across all reports (2M abstracts) with case reports containing numeric information (the ~900k we sample from), we see a massive reduction ($>30\%$) in terms associated with the following topics: surgery and post-surgery care, dentistry, ophthalmology, prostheses and rehab, patient care and nursing, some mental disorders and circulatory diseases/issues. Hence, we expect these topics to be relatively undersampled in our pool of case reports.

\noindent
\textbf{Annotation Pilots:} During pilots, we also conducted one or more disagreement discussion sessions per pilot round. These discussions were helpful in providing annotators the opportunity to highlight important spans/relations being missed by the schema, which led to the addition of the subpopulation entity, demographic attribute, and subpopulationof and treatmentof relations. Despite the introduction of some new elements, inter-annotator agreement continued to increase steadily over the pilot rounds, as shown in Table~\ref{tab:roundagreement} before plateauing at the end of round 3.

\noindent
\textbf{Full-Scale Annotation:} During rounds 1-3 of full-scale annotation, annotators were provided batches of 25 abstracts each. As their familiarity with the annotation schema and ability to handle ambiguous cases improved, we provided larger batches of 100 abstracts each during rounds 4-6. After each round, agreement was assessed and disagreement dicussions were conducted to discuss ambiguous cases, if needed, which ensured that agreement was maintained across rounds as seen from Table~\ref{tab:roundagreement}. Tables~\ref{tab:entityagreement},~\ref{tab:attributeagreement} and~\ref{tab:relationagreement} present final agreement scores per entity type, attribute type and relation type respectively. From these tables, we can see that Subpopulation and Intervention entities are the trickiest to annotate, leading to lower agreement on SubpopulationOf and InterventionOf relation types due to error cascading (i.e., if entity annotations don't match, relation annotations are unlikely to match either).

\section{Inter-Annotator Agreement}
Table~\ref{tab:roundagreement} shows the evolution in inter-annotator agreement over our initial pilot rounds, as well as the level of inter-annotator agreement maintained during each round of the full-scale annotation process. We see a large increase in relation agreement from pilot 1 to pilot 2, and consistent agreement scores across all tasks in all rounds thereafter. Tables~\ref{tab:entityagreement},~\ref{tab:attributeagreement} and~\ref{tab:relationagreement} present inter-annotator agreement breakdown according to entity, attribute and relation types in our schema.



\section{Hyperparameter Details}
\label{sec:hyperparam}
\noindent
\textbf{Extractive Models:}
\begin{itemize}[leftmargin=*,topsep=0pt]
\setlength\itemsep{-0.5em}
\item \textbf{OneIE:} We use an overall learning rate and weight decay of $1e-3$, and a learning rate and weight decay of $1e-5$ for the BERT component, a batch size of 10, and gradient clipping value of $5.0$. The model is trained for 60 epochs with a 5-epoch warmup phase.
\item \textbf{PURE:} We use a context window size of 300 words, overall learning rate of $1e-5$, task learning rate of $5e-4$, batch size of 16, and train for 100 epochs.
\item \textbf{LocLabel:} We use a learning rate of $5e-6$, warmup rate of $0.1$, weight decay of $0.01$, gradient clipping value of $1.0$, batch size of $6$ and train for 35 epochs. LocLabel also requires word vectors, for which we use the 200-dimensional Pubmed-trained word2vec embeddings (BioWordVec) released by \citet{zhang2019biowordvec}, which are available at \url{https://github.com/ncbi-nlp/BioWordVec}.
\item \textbf{W2NER:} We use an overall learning rate of $1e-3$ and a learning rate of $5e-6$ for the BERT component, no weight decay, warmup factor fo $0.1$, gradient clipping value of $5.0$, batch size of $8$, and train for 10 epochs.
\end{itemize}

\noindent
\textbf{Generative Models:} All models are trained for 10 epochs with a learning rate of $1e-5$, input context length of 1024, output length of 128, and a batch size of 2.

\noindent
\textbf{GPT3.5/GPT4:} We test the 16k and 8k context length versions of GPT3.5 and GPT4 respectively since our extraction tasks are abstract-level and require longer input contexts. We use the June 2023 versions of both models due to their \emph{function calling} capabilities, which leverage a structured JSON output format to improve information extraction capabilities. All experiments are run with a temperature of 0 and max output length of 512 tokens.

\section{Computing Infrastructure}
All LLM experiments are carried out on NVIDIA RTX A6000 GPUs with 48 GB RAM. Each finetuning run (FLAN-T5, BioGPT, BioMedLM) requires two GPUs with runtimes ranging from 9-17 hours depending on task size and model size. We use the DeepSpeed integration from Huggingface, with ZeRO-3 optimization, for multi-GPU training.

\section{Prompt Details}
\label{sec:prompts}
\begin{figure*}
    \includegraphics[scale=0.63]{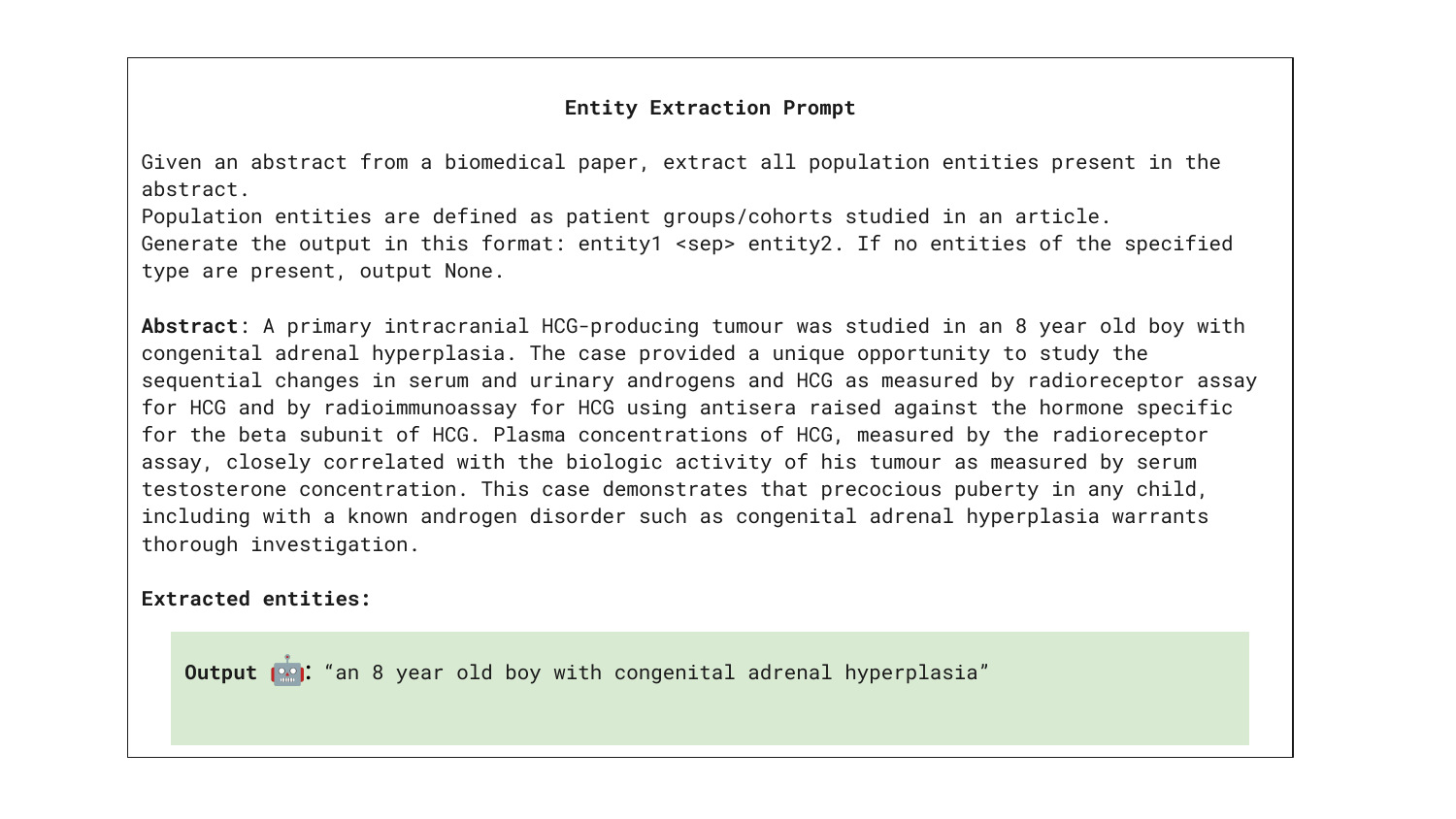}
    \caption{Example prompt used to evaluate the performance of finetuned LLMs on entity extraction. Such prompts are generated for all seven entity types in our dataset.}
    \label{fig:flan_entity}
\end{figure*}

\begin{figure*}
    \includegraphics[scale=0.63]{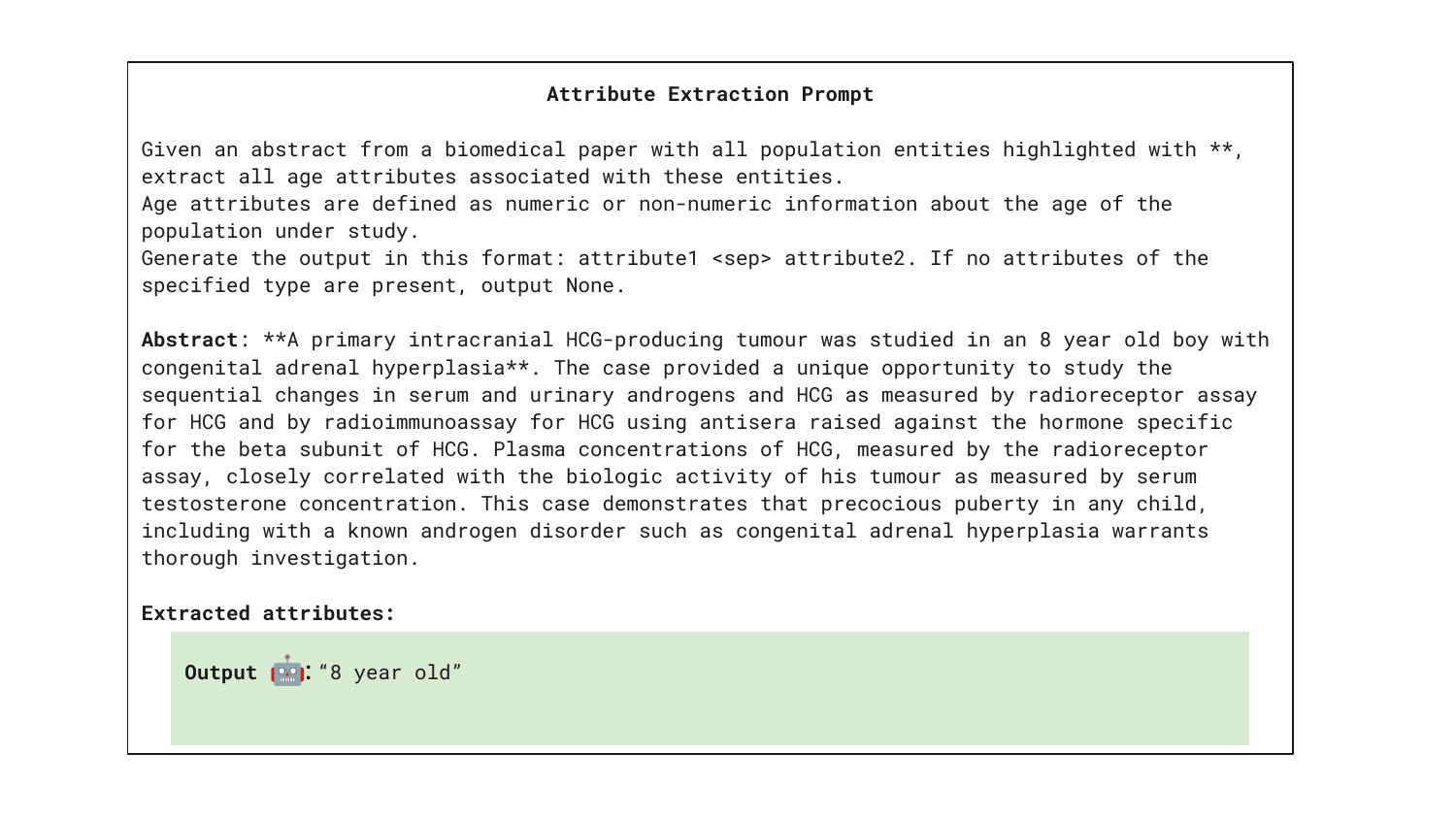}
    \caption{Example prompt used to evaluate the performance of finetuned LLMs on attribute extraction with gold entities provided. Such prompts are generated for all nine attribute types in our dataset.}
    \label{fig:flan_attribute}
\end{figure*}

\begin{figure*}
    \includegraphics[scale=0.63]{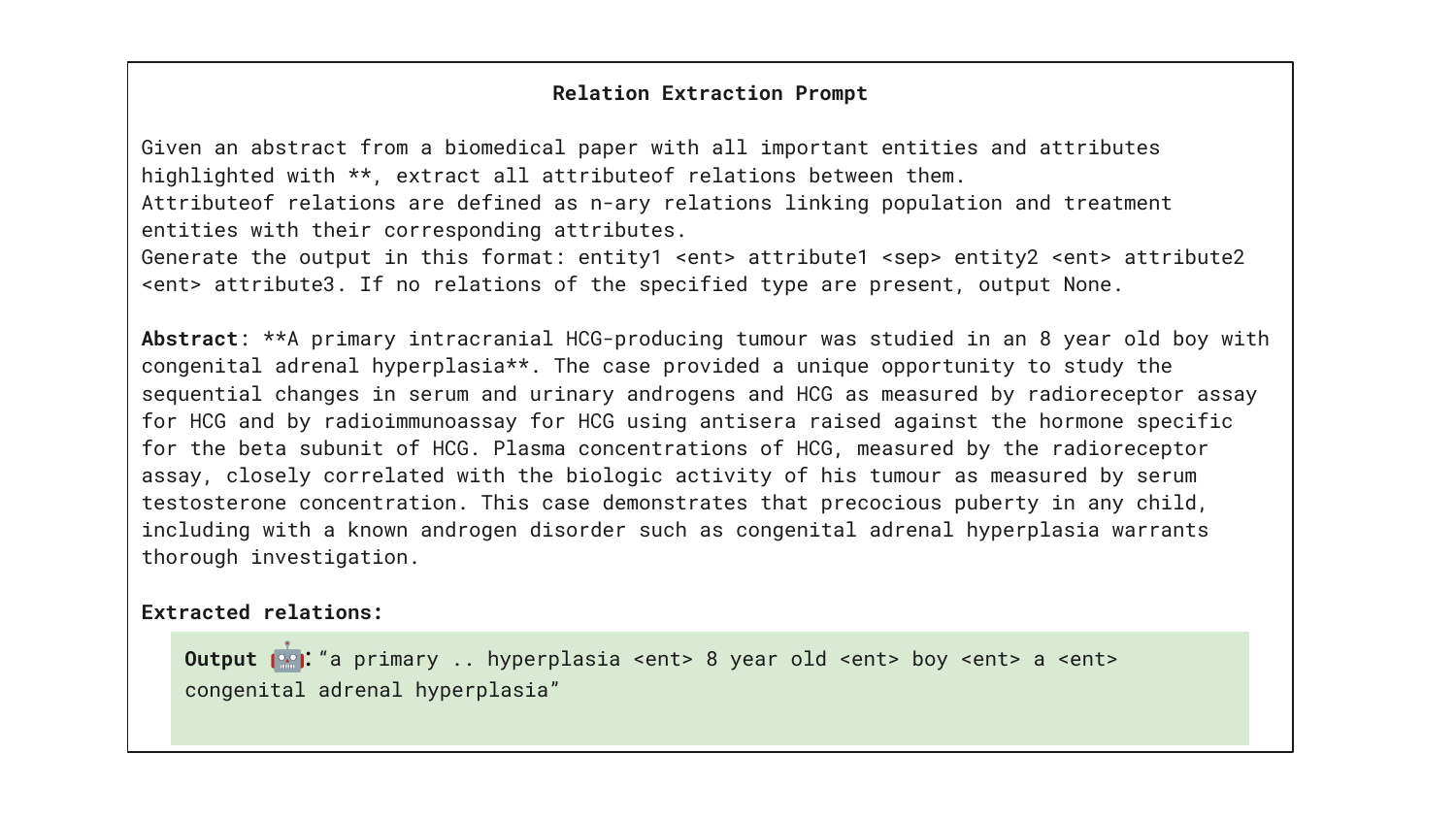}
    \caption{Example prompt used to evaluate the performance of finetuned LLMs on relation extraction with gold entities and attributes provided. Such prompts are generated for all four relation types in our dataset.}
    \label{fig:flan_relation}
\end{figure*}

\begin{figure*}
    \includegraphics[scale=0.63]{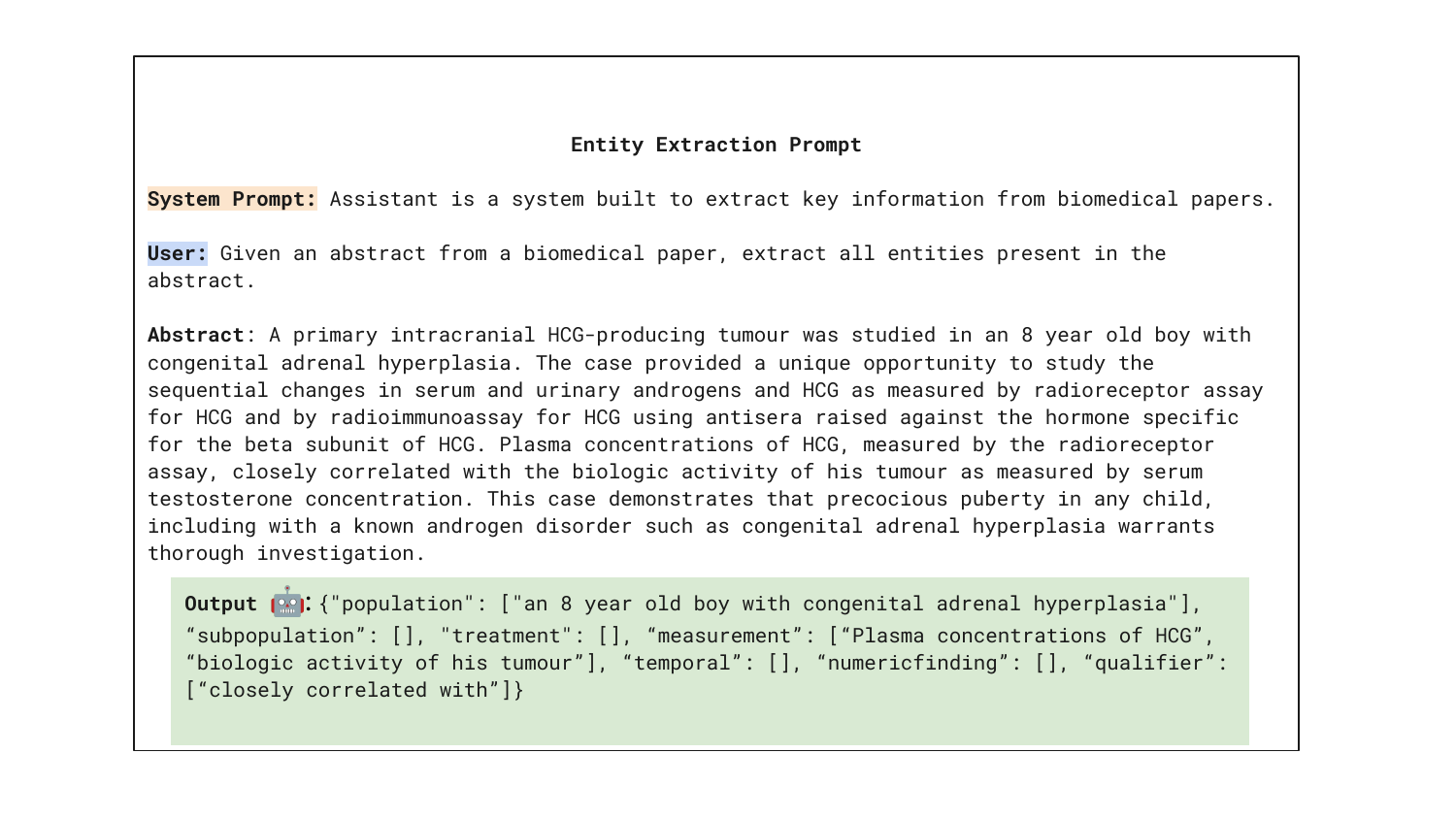}
    \caption{Prompt used to evaluate the performance of GPT-3.5 and GPT-4 on the entity extraction task. Additionally, entity type definitions from Table~\ref{tab:entity_types} are provided as \textit{function parameters} to leverage OpenAI's function calling capabilities.}
    \label{fig:gpt_entity}
\end{figure*}

\begin{figure*}
    \includegraphics[scale=0.63]{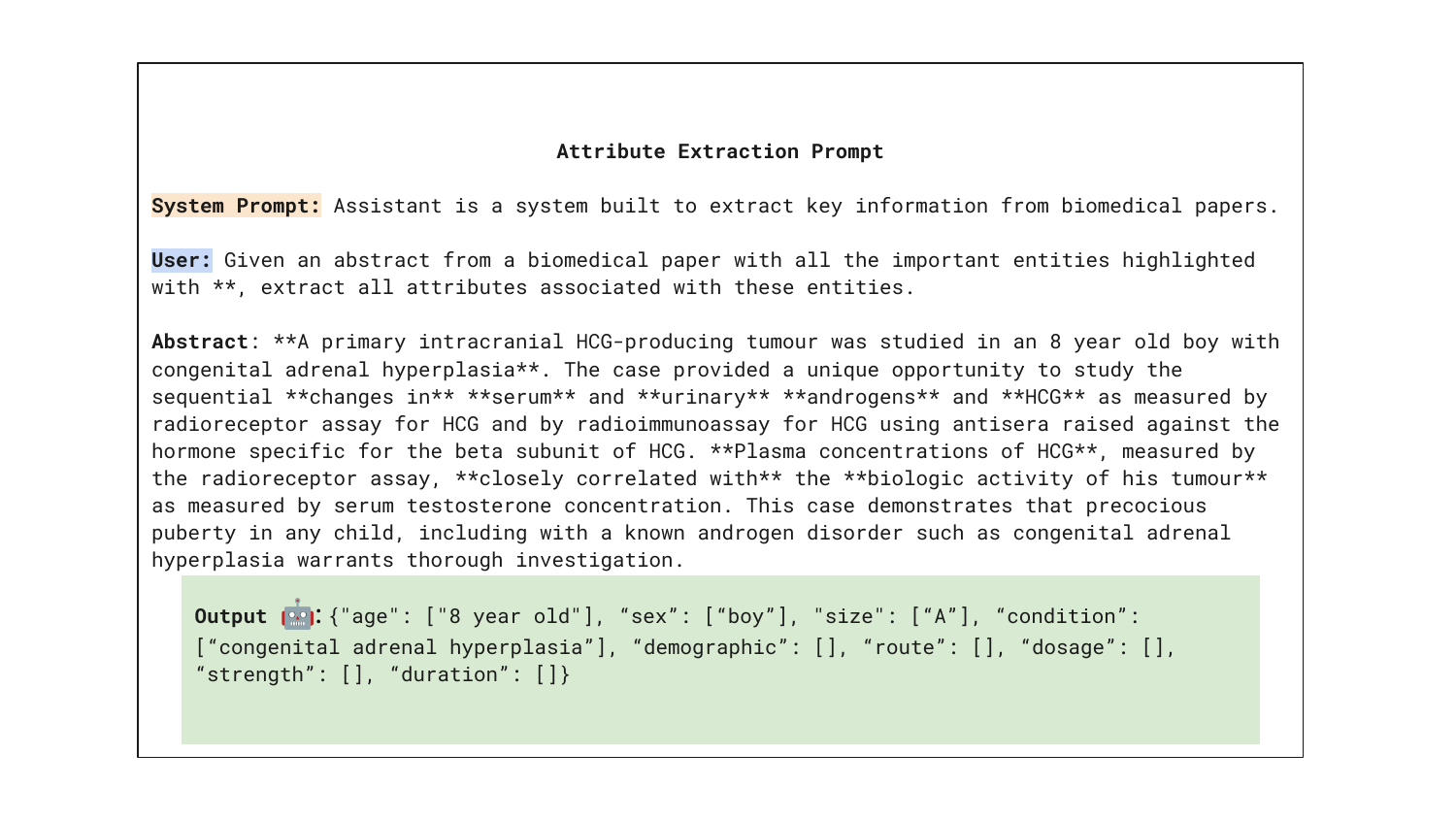}
    \caption{Prompt used to evaluate performance of GPT-3.5 and GPT-4 on the attribute extraction task with gold entities provided. Additionally, attribute type definitions from Table~\ref{tab:attribute_types} are provided as \textit{function parameters} to leverage OpenAI's function calling capabilities.}
    \label{fig:gpt_attribute}
\end{figure*}

\begin{figure*}
    \includegraphics[scale=0.63]{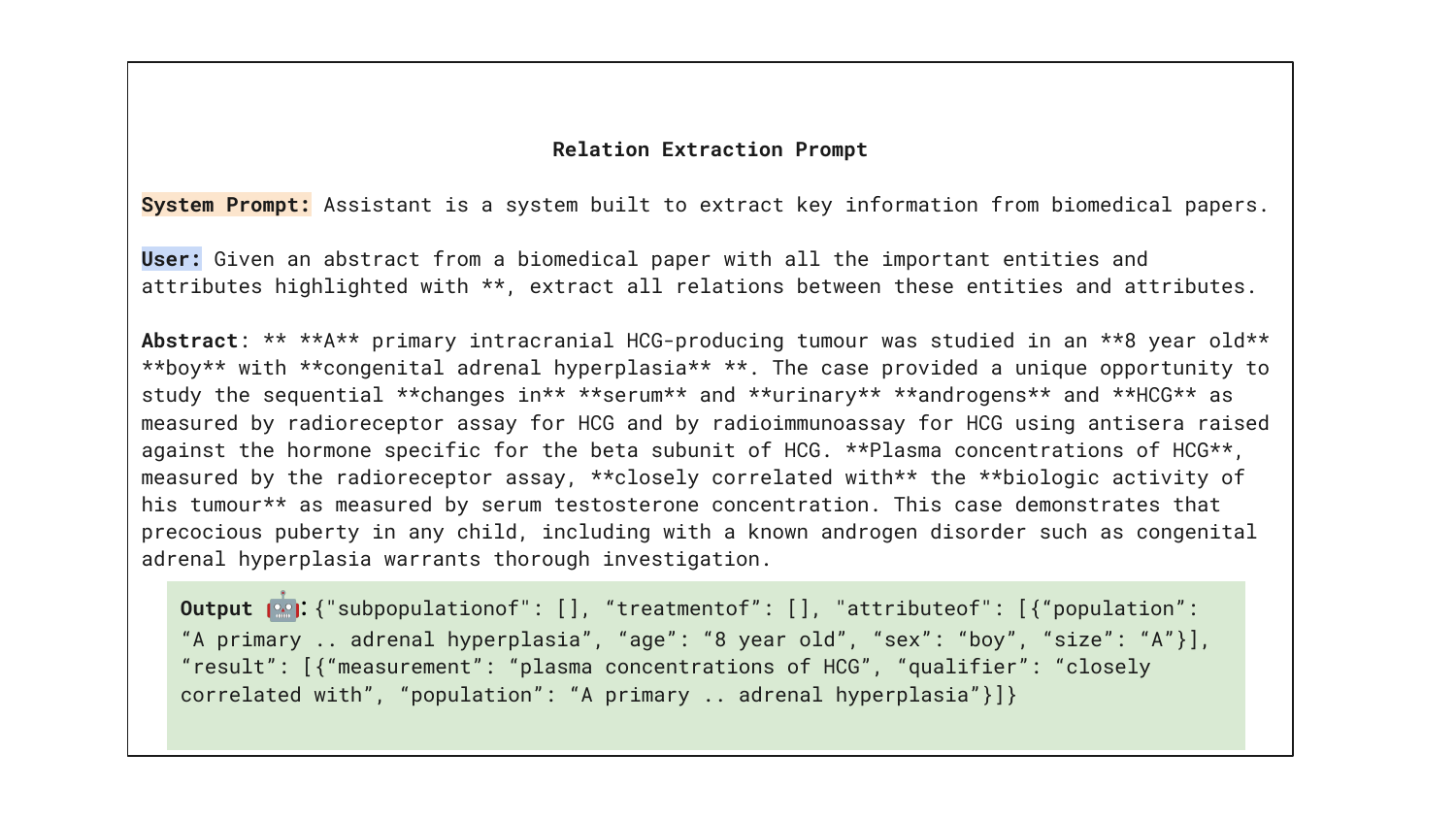}
    \caption{Prompt used to evaluate performance of GPT-3.5 and GPT-4 on the relation extraction task with gold entities and attributes provided. Additionally, relation type definitions from Table~\ref{tab:relation_types} are provided as \textit{function parameters} to leverage OpenAI's function calling capabilities.}
    \label{fig:gpt_relation}
\end{figure*}

Figures~\ref{fig:flan_entity},~\ref{fig:flan_attribute} and~\ref{fig:flan_relation} present the prompts used to evaluate the performance of finetuned LLMs (FLAN-T5, BioGPT and BioMedLM) on entity, attribute and relation extraction respectively. Similarly, Figures~\ref{fig:gpt_entity},~\ref{fig:gpt_attribute} and~\ref{fig:gpt_relation} present the prompts used to evaluate the performance of GPT-3.5 and GPT-4 models (in a zero-shot setting) on entity, attribute and relation extraction respectively. For the in-context learning setting, additional few-shot examples are appended to the prompt before providing the abstract.

\end{document}